\def\year{2022}\relax

\newcommand{\mytitle}{MACQ: A Holistic View of Model Acquisition Techniques}

\documentclass[letterpaper]{article} % DO NOT CHANGE THIS
\usepackage{aaai22}  % DO NOT CHANGE THIS

\usepackage{times}  % DO NOT CHANGE THIS
\usepackage{helvet} % DO NOT CHANGE THIS
\usepackage{courier}  % DO NOT CHANGE THIS
\usepackage[hyphens]{url}  % DO NOT CHANGE THIS
\usepackage{graphicx} % DO NOT CHANGE THIS

\usepackage{xspace}
\usepackage[colorlinks=true, linkcolor=blue, urlcolor=blue, citecolor=blue]{hyperref}

\usepackage{natbib}  % DO NOT CHANGE THIS AND DO NOT ADD ANY OPTIONS TO IT
\usepackage{caption} % DO NOT CHANGE THIS AND DO NOT ADD ANY OPTIONS TO IT
\usepackage[frozencache,cachedir=.]{minted}

\nocopyright

\title{\mytitle}

\author {
    Ethan Callanan\equalcontrib\textsuperscript{\rm 1}\hspace{2mm}
    Rebecca De Venezia\equalcontrib\textsuperscript{\rm 1}\hspace{2mm}
    Victoria Armstrong\textsuperscript{\rm 1}\hspace{2mm}
    Alison Paredes\textsuperscript{\rm 1}\\
    Tathagata Chakraborti\textsuperscript{\rm 2}\hspace{2mm}
    Christian Muise\textsuperscript{\rm 1}\\[1ex]
}
\affiliations {
    \textsuperscript{\rm 1} Queen's University, 
    \textsuperscript{\rm 2} IBM Research AI \\
    Contact: \texttt{christian.muise@queensu.ca}
}

\newcommand{\macq}{\texttt{MACQ}\xspace}
\newcommand{\tarski}{\textsc{Tarski}\xspace}

\begin{document}

    \maketitle

\begin{abstract}
For over three decades, the planning community has explored countless methods for data-driven model acquisition. These range in \emph{sophistication} (e.g., simple set operations to full-blown reformulations), \emph{methodology} (e.g., logic-based -vs- planing-based), and \emph{assumptions} (e.g., fully -vs- partially observable). With no fewer than 43 publications in the space, it can be overwhelming to understand what approach could or should be applied in a new setting. We present a holistic characterization of the action model acquisition space and further introduce a unifying framework for automated action model acquisition. We have re-implemented some of the landmark approaches in the area, and our characterization of all the techniques offers deep insight into the research opportunities that remain; i.e., those settings where no technique is capable of solving.

\vspace{5pt}
{\bf Project}: \href{macq.planning.domains}{macq.planning.domains}

% {\bf GitHub}: \href{https://github.com/AI-Planning/macq}{https://github.com/QuMuLab/macq}

% {\bf Visualize:} \href{http://ibm.biz/macqviz}{ibm.biz/macqviz}
\end{abstract}
    
\section{Introduction}
\label{sec:introduction}

Model acquisition has been a hallmark sub-field of automated planning for decades. From the early approaches to extracting simple planning models in an automated way \cite{shen1989rule} to modern techniques for extracting lifted action theories from state tokens alone \cite{abs-2105-10830}, the planning community has developed dozens of approaches to model acquisition for various settings. Simultaneously, there have been a few surveys that capture the status of this 
rich sub-field and characterize the approach across many axes 
\cite{JimenezRFFB12, hud20380, AroraFPMP18}.
What is missing from the collective focus on the suite of techniques available is a unifying framework that allows researchers to explore both the existing techniques, as well as the gaps in what we are capable of solving.
While the explosion in model learning techniques has dated past surveys 
pretty quickly, our open-source approach and live interface to the 
repository aims to make \macq the one-stop-shop for learning planning models.

In this work, we present the first major step towards such a unified framework. Both from a theoretical standpoint -- characterizing model acquisition techniques in terms familiar to the planning community -- and from the practical standpoint with growing implementation of the core algorithms. Our contribution goes substantially further than just a survey of the existing techniques. By integrating them under a single theoretical framework and implementation, we effectively open the door to systematically characterizing the entire sub-field of research.

We accomplish this characterization by identifying the key properties that distinguish model acquisition techniques and then applying knowledge compilation techniques to map out what is known from the field. We are thus able to holistically view what is possible theoretically (i.e., papers exist to address the setting), what is possible practically (an implementation exists in the framework), and what remains an open research question.

Finally, having the suite of tools for state trace generation, modification, and analysis all under a single framework allows us to rapidly test different approaches to a specific setting. New use-cases for model acquisition can leverage the growing library of implemented techniques for model acquisition, and this provides tangible benefits to those outside the planning community who wish to apply planning techniques in their domain of expertise.

%In the following section, we introduce the background necessary to characterize the techniques we consider and then
% We present our core methodology in Section \ref{sec:methodology} and elaborate on the compilation of research recommendations in Section \ref{sec:research-rec}. In Section \ref{sec:demo}, we demonstrate the range of functionality with the project and its interface. 
% and follow with a discussion of related work in Section \ref{sec:related}. 
% Finally, we wrap with a 
% general discussion in Section \ref{sec:discussion} and 
% dedicate Section \ref{sec:open-questions} 
% to the open research questions identified both by the authors 
% as well as the system's own analysis of existing literature.

    % \input{section/2-background}
    \section{\macq Framework}
\label{sec:methodology}

There are three core components to the \macq framework: (1) trace generation; (2) observation tokenization; and (3) model extraction. Not all of these are mandatory, but each complements the others to offer a rich array of functionality. We discuss each of them in turn in Section \ref{sec:methodology}, 
but lead this section with a discussion of key features and assumptions
of concern in describing a model learning task.

% we may find with planning traces -- both in the assumptions on the underlying model and in the data representation itself. 

\subsection{Feature Analysis}
\label{subsec:features}

A model learning task involves three key considerations -- 
1) what are the features of the agent whose model is being learned; 
2) what are the features of the model being learned; and
3) what are the features of the data from which the model is being learned.
Note the distinction between (2) and (3). This separation of features
allows our framework with the flexibility to
provide the user with the choice of what 
kind of model to learn from what kind of data. 
For example, a user can choose to use a stochastic model extraction technique 
on noisy data instead of modelling noise directly. More on this
``token casting'' feature later in Section \ref{subsubsec:token-casting}.

\subsubsection{Agent Features}
\begin{itemize}
\item {\bf Rationality} The primary consideration here is whether the observed
agent is rational or not. By default, we assume that this is unknown or 
that there is no assumption of rationality in an extraction technique.
If there is, we allow for two types of rationality -- 
optimal traces (in the classical
sense) or causally relevant traces where there are no redundant actions
in a plan i.e. there are no (subsets of) actions that 
can be removed from the trace and the agent can still reach its goal.
\end{itemize}

\subsubsection{Model Features}
\begin{itemize}
    \item {\bf Uncertainty} The first feature of interest is 
    whether uncertainty is captured by the model -- a model
    can take three forms: 1) deterministic, 2) non-deterministic, 
    and 3) probabilistic. Note that (3) implies (2)and (1);
    and (2) implies (1). Thus, a model can have at least one and at most one
    of these features -- this is generally not true for the rest of the features.
    \item {\bf Parameterization} These features capture whether actions (as well
    as predicates) in the model are parameterized by the objects they operate on
    and whether those objects in turn are typed or atomic.
\end{itemize}

\subsubsection{Data Features}
\begin{itemize}
    \item {\bf Observability} One of the primary features of concern in a 
    model learning task is how much of the environment is observable.
    The fluents describing the state may be 1) fully or 2) partially observable 
    or 3) not observable at all. Similar to the discussion on model uncertainty,
    the ability of a model extraction technique to deal with (2) implies (1); 
    but (3) implies neither of (1) or (2).
    \item {\bf Action Observability} In addition to whether the fluents are 
    observable, an additional consideration is whether the action labels and 
    parameters are known and whether there is a seed model, to begin with.
    \item {\bf Parameterization} This mirrors the model features by the same names.
    Additionally, it also includes the possibility to have action costs.
    \item {\bf Noise} Noise in data may manifest either as noise in fluent observations or noise in observed action labels.
    \item {\bf Access} to the initial and goal state of a trace.
    \item {\bf Trace} Finally trace-level features include whether there 
    is access to the cost of the plan and to what extent (full or partial) the trace
    is observable i.e. whether there are missing actions or not.
\end{itemize}

\subsection{Trace Generation}
Plan traces may be sourced from a variety of sources, and the first component (trace generation) serves as both a rich set of techniques to \textit{generate} planning traces as well as a suite of tools to parse and process existing data for model extraction. Here, we describe some of the highlights in functionality for the ``trace generation'' part of the \macq ecosystem.

\subsubsection*{CSV Processing}
As a trivial baseline for trace generation, the \macq library offers functionality to load and package up simple CSV files. Columns with the values \verb|0| or \verb|1| will be retained (presumed to be boolean fluents), and a single specially designated column for the ``action'' label is required. This does not cover the full set of data format assumptions (e.g., parameterized actions), but is nonetheless a common starting point for many model acquisition tasks.

\subsubsection*{Statespace Sampling}
Given any valid classical planning problem, \macq has the ability to generate random statespace trajectories. The PDDL model supplied (either as raw PDDL, pointers to existing files, or a reference to 
the \href{https://api.planning.domains/}{api.planning.domains} problem ID \cite{planning.domains}) is parsed by the \tarski library \cite{tarski}, and actions are selected (uniformly at random) starting in the initial state. The goal is ignored in this case, and \macq will generate the predefined number of traces at a given length.
As an added feature, we have also incorporated the methodology adopted by the FastDownward system \cite{helmert2006fast} to perform random state sampling to a depth based on heuristic computation.\footnote{Details: \url{https://github.com/aibasel/downward/blob/main/src/search/task_utils/sampling.cc}}

\subsubsection*{Goal-Oriented Sampling}
Another approach for generating a single trace is to compute a plan for the given domain/problem pair. However, as a generation technique, it is limited to generating just a single trace. \macq has expanded on this by allowing for random goal sampling. The approach works as follows:

\begin{enumerate}
    \item Use the statespace sampling to compute a sequence of actions/states of length $k$.
    \item Sample subsets of fluents of size $g$ from the final state in this sequence such that,
    \begin{enumerate}
        \item It reflects the same type of fluents in the original goal.
        \item It is not easily achieved from the original initial state.
    \end{enumerate}
    \item Use those sampled subsets as goals for computing a plan.
\end{enumerate}

There are several design decisions to be made: user-selected values for $k$ and $g$ (which may be domain-specific); sampling procedure in step 2 to adhere to 2(a); measuring the quality of a goal candidate in step 2(b); etc. \macq currently includes a preference to use fluents corresponding to the goal predicates for step 2(a), and uses a computed plan as a proxy for 2(b) -- the closer the found plan is to length $k$, then the better the goal candidate is.
This technique provides a rich mechanism for data generation given a single seed planning problem. Not only is it useful for exploring model extraction techniques, but we predict it may have wider use in the area of planning and learning.

\subsection{Observation Tokenization}
There is a vast array of model acquisition techniques that exist (some surveys on the space are discussed in Section \ref{sec:introduction}). In an effort to provide a common foundation for a library that encompasses this rich variety, we appeal to the notion of \emph{observation tokens} \cite{2013Geffner}. To unify various approaches for planning with partial observability, \citeauthor{2013Geffner} introduces a notion of an observation token. It succinctly captures what the agent sees and can act upon. At times, this may be an indication of a sensing outcome. At other times, it may capture the partial state information viewable by the agent. We adopt this concept wholeheartedly for use in \macq.

\begin{quote}
    Every extraction method works on a set of observation token lists of a specific token type.
\end{quote}

\subsubsection{Observation Token Types}
Driven by the variety of extraction methods captured by \macq, we have identified several useful forms of observation tokens. These include:

\begin{itemize}
    \item \textsc{Identity}: Full state / action information is provided.
    \item \textsc{PartialState}: Some fluents are masked as unknown.
    \item \textsc{State ID}: No action or fluent information is provided -- same states correspond to the same token.
    \item \textsc{NoisyState}: Some fluents may be incorrectly assigned.
    \item \textsc{ActionOnly}: No state information is provided.
\end{itemize}

This list, although not exhaustive, provides a sense of the variety embedded within \macq. Each extraction method will designate the token types it is capable of processing, and this allows for the automatic discovery of methods applicable to data of a particular form. This means that \macq has the ability to automatically suggest the extraction methods that can work with a particular source of trace data.

\subsubsection{Tokenization}
To explore the effectiveness of extraction techniques, every form of observation token has the functionality to ``tokenize'' ground-truth data. For example, \textsc{PartialState} tokens can be created by specifying the likelihood of masking a fluent, along with the set of fluents that are eligible (defaulting to the entire state).

This functionality is essential for the development of new extraction methods, as well as testing pre-existing ones. Combined with the methods for trace generation in \macq, this provides a robust means for data generation in the model acquisition space.

\subsubsection{Token Casting}
\label{subsubsec:token-casting}

The extraction techniques are tightly coupled with the type of representations they can handle. An approach for partially observable states will only operate on the appropriate class of \verb|PartiallyObservable| token types for the trace. To extend this, we allow for ``token casting'', which will (if possible) transform tokens of one type to another. Taking our example further, if we have fully observable states and we wanted to test a technique implemented in \macq dedicated to partially observable state spaces (because of other functionality it offers), we can ``token cast'' the trace data to partially observable tokens.

Token casting is not always feasible, but the \macq framework is set up such that finding these casting paths is naturally available. Every contribution to the space of model acquisition is characterized by the limited scope of the paper/work. Through methods like tokenization and token casting, we open the door to applying techniques in a richer variety of settings, not originally envisioned by the authors.

\subsection{Action Model Extraction}
The bulk of the \macq project is dedicated to the extraction of action theories. At the time of writing, we have re-implemented a representative sample of model acquisition techniques spanning several features discussed in Section \ref{subsec:features} to demonstrate the potential of the \macq system:

\begin{itemize}

    \item \textsc{Observer} \cite{observer}: One of the first and most simple methods for model acquisition, this technique assumes full observability, noise-free data, and deterministic actions. The extracted theories are in STRIPS form, and the methods are mostly set-based.
    
    \item \textsc{ARMS} \cite{arms,armsj}: This line of work handles partially observable states (fluents are hidden) as well as traces (entire states may be missing). There is an assumption that the goal and initial states are known, and while the actions and predicates are parameterized, they are not typed. The technique uses MaxSAT to solve a particular encoding for model extraction.
    
    \item \textsc{SLAF} \cite{slaf}: This method also focuses on partially observable state traces and appeals to a logical encoding to extract the action theories. It is capable of producing one action theory, or several that fit the data. It operates by iteratively calling a SAT solver to find the entailments that lead to a valid theory (i.e., one that adheres to the observations). The input traces are assumed to be noise-free.
    
    \item \textsc{AMDN} \cite{amdn}: This technique relies on MaxSAT to find the most likely action models given potentially disordered and noisy plan traces (here, noise means actions may be out of order). Further, the states may be partially observable and noisy, and the actions occur in parallel.
    
\end{itemize}

A common element of these approaches is the heavy reliance on SAT or MaxSAT technology. Because of this commonality, several elements of functionality have been included in the \macq project for extraction techniques to make use of. These include model building, solving with wrapped binaries, and solution extraction. Specifically, the project relies on the Bauhaus \cite{bauhaus}, python-nnf \cite{pynnf}, and PySAT \cite{pysat} libraries, as well as the kissat \cite{kissat} and RC2 \cite{rc2} SAT / MaxSAT solvers. While the current focus is SAT-based, the full scope of model acquisition techniques are planned for eventual implementation.

Finally, custom and flexible representations for learned actions and fluents are shared across the approaches. This allows for a uniform treatment of what is produced -- regardless of extraction technology -- and further allows us to efficiently generalize the serialization of the action theories. This final step (writing to PDDL) is achieved using the Tarski library \cite{tarski}.

    \section{\macq in Action}
\label{sec:demo}
Here, we briefly showcase some of the interface for the toolkit, as well as the summary view of the work.

\begin{figure}[tbp!]
\scriptsize
\begin{minted}[mathescape,
               linenos,
               numbersep=5pt,
               gobble=0,
               frame=lines,
               framesep=2mm]{csharp}
               
from macq import generate, extract 
from macq.trace import PlanningObject, Fluent, TraceList 
from macq.observation import PartialObservation 

def get_fluent(name: str, objs: list[str]): 
    objects = [PlanningObject(o.split0[0], o.split()[1]) 
               for o in objs] 
    return Fluent(name, objects) 

traces = TraceList() 
generator = generate.pddl.TraceFromGoal(problem_id.1801) 

generator.change_goal({ 
    get_fluent(
        "communicated_soil_data", 
        ["waypoint waypoint2"]
    ),
    get_fluent(
        "communicated_rock_data", 
        ["waypoint waypoint3"]
    ),
    get_fluent(
        "communicated_image_data", 
        ["objective objective)", "mode high_res"] 
    ), 
})
traces.append(generator.generate_trace()) 

generator.change_goal({ 
    get_fluent(
        "communicated_soil_data", 
        ["waypoint waypoint3"]
    ),
    get_fluent(
        "communicated_rock_data", 
        ["waypoint waypoint2"]
    ),
    get_fluent(
        "communicated_image_data", 
        ["objective objective)", "mode high_res"]
    ),
}) 
traces.append(generator.generate_trace()) 

observations = traces.tokenize(
    PartialObservation, 
    percent_missing = 0.60
) 
model = extract.Extract(
    observations, 
    extract.modes.ARMS, 
    upper_bound = 2, 
    min_support = 2, 
    action_weight = 110, 
    info_weight = 100, 
    threshold = 0.6, 
    info3_default = 30, 
    plan_default = 30, 
)
print(model.details())
\end{minted}
\caption{\macq usage}
\label{fig:arm-example}
\end{figure}

\begin{figure}[!th]
\scriptsize
\begin{verbatim}
Actions:
(communicate_soil_data waypoint
                       lander
                       rover
                       waypoint):
  precond:
    at rover waypoint
  add:
    at_rock_sample waypoint
    have_rock_analysis rover waypoint
    communicated_soil_data waypoint
    channel_free lander
    at_soil_sample waypoint
  delete:
(communicate_image_data lander
                        waypoint
                        rover
                        objective
                        mode
                        waypoint):
  precond:
    calibrated camera rover
    communicated_rock_data waypoint
    at_rock_sample waypoint
    have_soil_analysis rover waypoint
    channel_free lander
    at rover waypoint
  add:
    have_image rover objective mode
    calibrated camera rover
    communicated_image_data objective mode
  delete:
    calibrated camera rover
(drop store rover):
  precond:
    have_image rover objective mode
    have_soil_analysis rover waypoint
    available rover
    calibrated camera rover
    at rover waypoint
  add:
    have_image rover objective mode
    calibrated camera rover
  delete:

 ...
\end{verbatim}
\caption{Sample of the output from the ARMS extraction.}
\label{fig:arm-example-output}
\end{figure}

\subsection{Library Usage}
Corresponding to the three main components detailed in Section \ref{sec:methodology} -- trace generation, observation tokenization, and model extraction -- the \macq library offers a range of functionality for each. Usage of \macq also follows this natural order of first generating or loading traces, then optionally applying tokenization, and then doing the model extraction.

Figure \ref{fig:arm-example} shows the code required to (1) generate traces for a problem found in the online repository at \url{api.planning.domains}, (2) tokenize by removing 60\% of the fluents seen, and (3) apply the ARMS algorithm to extract potential actions. A portion of the output is shown in Figure \ref{fig:arm-example-output}.

As long as the type of tokens in a trace allows for it\footnote{See the Section on ``Token Casting'' for details on how traces can be transformed to different types.}, various extraction techniques can be substituted and compared. Similarly, various data sources (from generative to pre-existing) can be used to seed the entire approach.

The \macq library was built from the ground up to be (1) extensible and generalizable to all of the common model acquisition techniques; (2) serve as a rich resource for practitioners looking to apply model acquisition; and (3) provide a foundation for new research in the area of model acquisition.

\begin{figure}
\centering
\includegraphics[width=\linewidth]{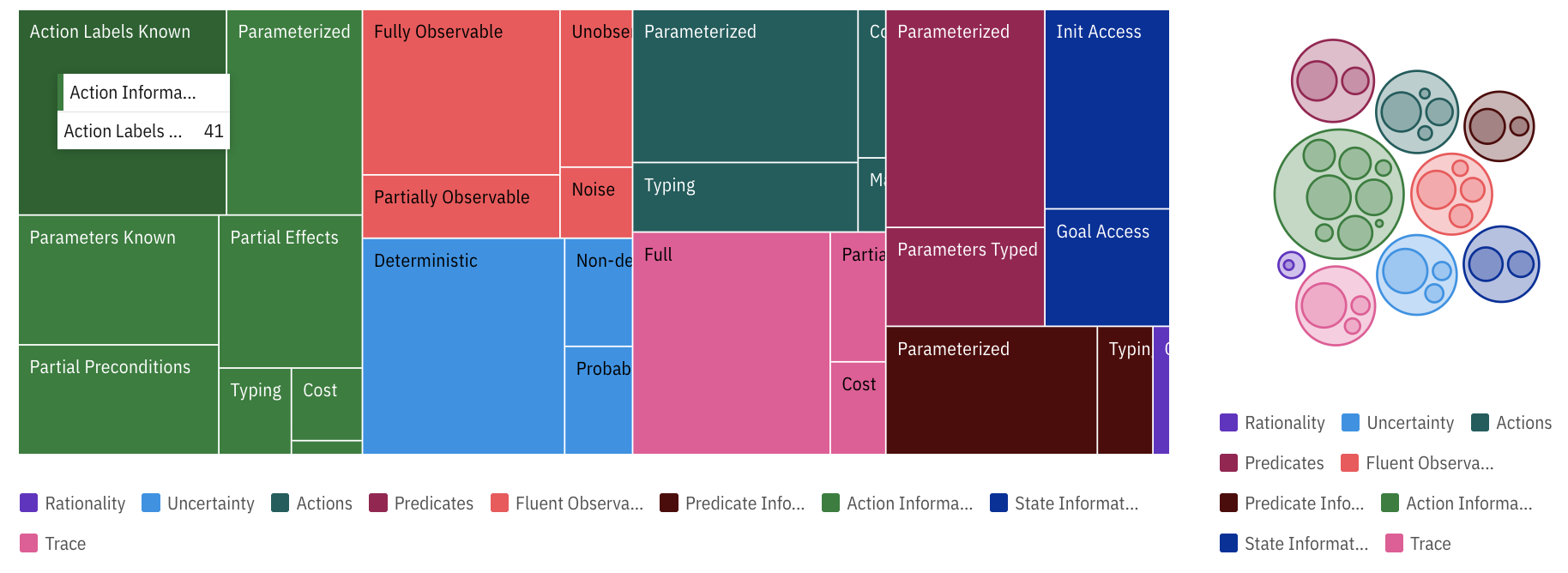}
\caption{\macq treemap view.}
\label{fig:treemap}
\end{figure}

\begin{figure}
\centering
\includegraphics[width=\linewidth]{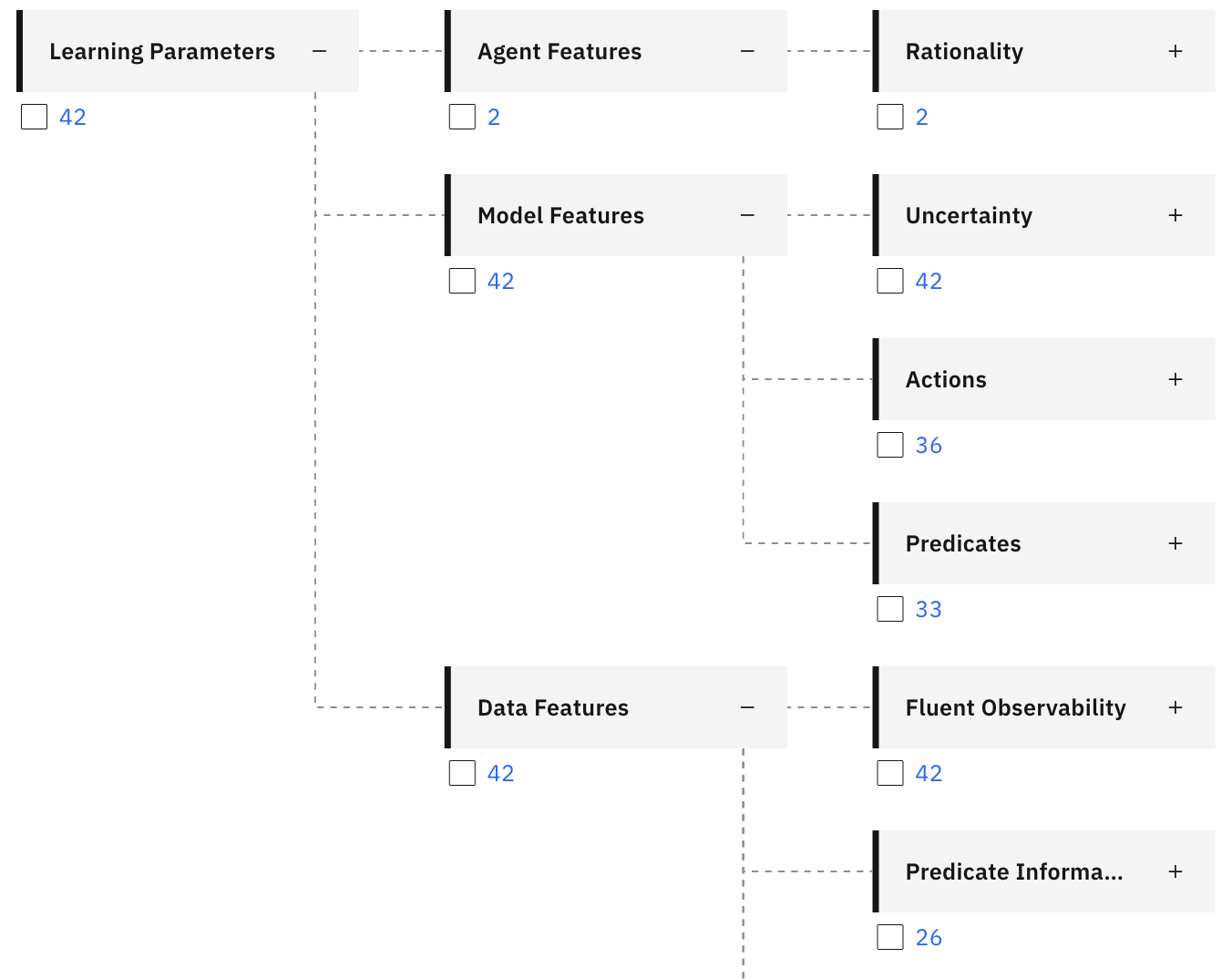}
\caption{\macq hierarchical view.}
\label{fig:hierarchy}
\end{figure}

\subsection{Visual Interface}
\label{subsec:viz}

The \macq library also comes with a visual interface 
for users to explore the available papers on the topic 
through various lenses.
The primary view provides a taxonomic account 
of the various topics identified in the field and how
papers are classified along those topics -- for us,
these topics correspond to the features discussed
in Section \ref{subsec:features}.
This is shown in Figures \ref{fig:treemap} and \ref{fig:hierarchy}.

\begin{figure}[tbp!]
\centering
\includegraphics[width=\linewidth]{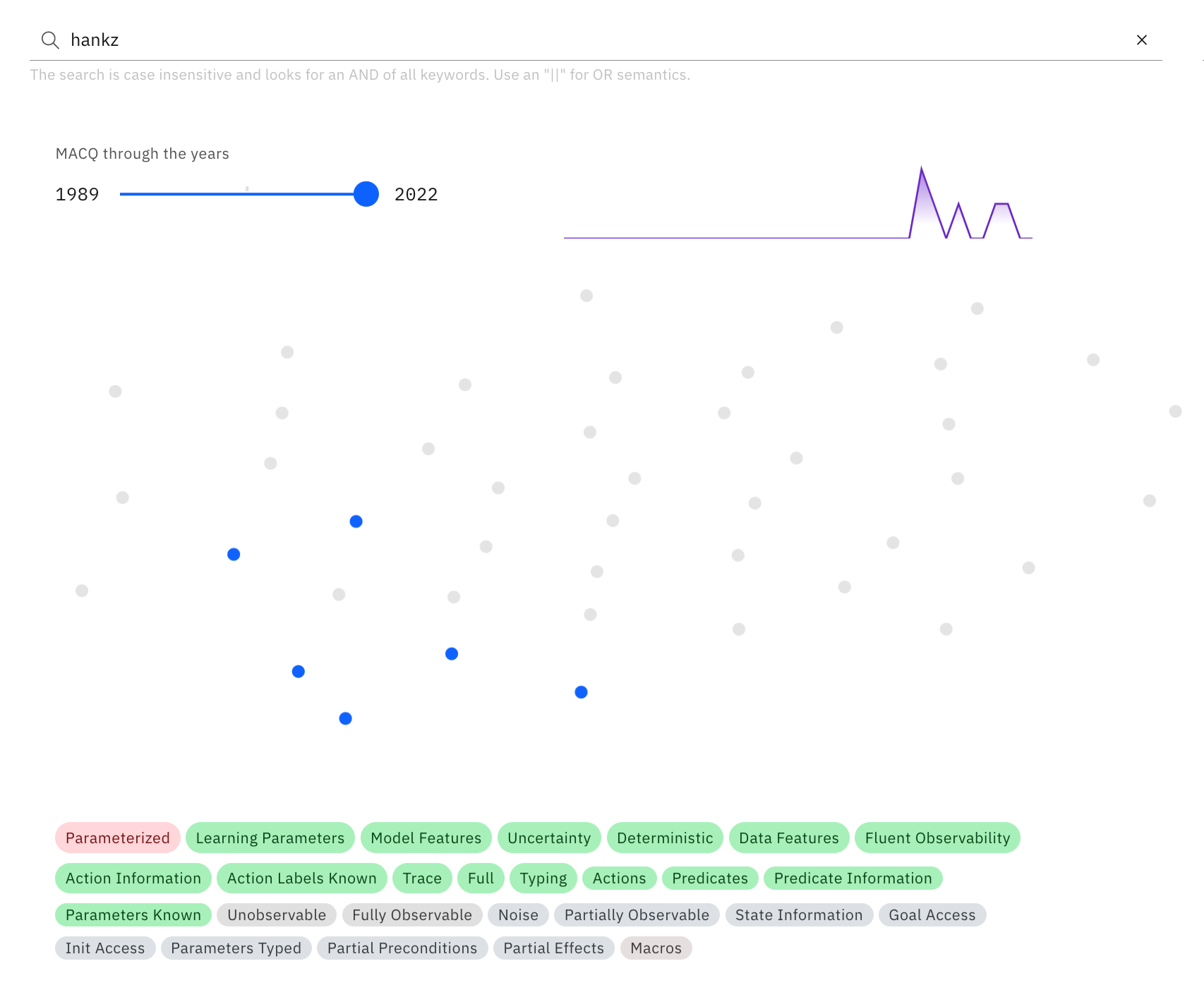}
\caption{Exploring papers written by Hankz Hankui Zhuo,
a prominent author in the field of model acquisition, in feature space.
Top right inset illustrates simulation of his over time while feature tags 
(sized and colourized by frequency) summarize salient topics in his papers.}
\label{fig:affinity}
\end{figure}

The next view displays the papers in \macq's knowledge
in the latent space of features. 
This document embedding is computed according to
the approach in \cite{specter2020cohan},
inspired by a similar application in \cite{RushStrobelt2020}.
In this view, the user can select subsets of papers in
feature space, filter by features by clicking on the tags
and even simulate the evolution of the feature space 
over time. Figure \ref{fig:affinity} provides an illustration
of the same.

\begin{figure}[tbp!]
\centering
\includegraphics[width=\linewidth]{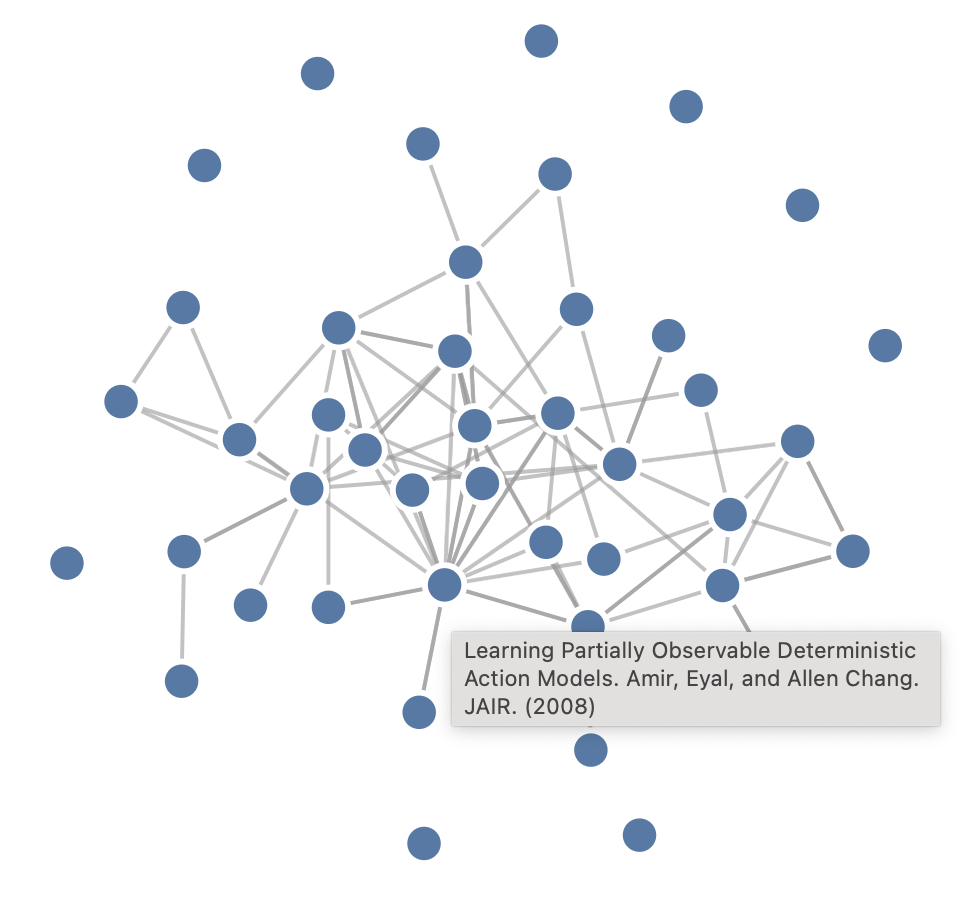}
\caption{Exploring the influence network in \macq -- this particular
cluster being hovered over belongs to \cite{slaf}, one of the seminal papers
in the field.}
\label{fig:network}
\end{figure}

Finally, from the PDF documents of the papers,
we also automatically extract a citation network 
to illustrate the most influential hubs in the world of \macq.
This is shown in Figure \ref{fig:network}.
As in the case of the similarity view, here too the user 
can modify the network view using the feature space 
as well as simulate how the network evolves over time.

    \section{Research Recommendations}
\label{sec:research-rec}
As a research field matures, our understanding of the gaps in our knowledge dwindles. Our efforts include not only a taxonomy of existing techniques for model acquisition (and an implementation of some of the most popular ones) but also a mechanism for exploring the research space as well. For every work documented by the \macq project, we have a feature vector that characterizes the technique. These are detailed above in Section \ref{subsec:features}. Further, we have a growing set of \textit{semantic constraints} over this taxonomy.

For example, ``if a technique can operate on partially observable traces, it must be able to work on fully observable traces'' and ``every technique must have full, partial, or no observability''. Specifying these constraints has one immediate benefit: it allows us to systematically verify the documented features of the existing approaches. This has led to several ``bug fixes'' of the data collected already. However, the true power lies in the potential for viewing the research field both holistically and logically.

Alongside \macq, and in collaboration with the visualization project used to exhibit the research area, we have developed a logical theory that corresponds to the \textit{valid space of research} according to the features detailed in Section \ref{subsec:features} and manually specified constraints over them. All of the model acquisition techniques are validated but further encoded as constraints themselves (their features being converted to a conjunction of Boolean variables or their negation).

\begin{figure*}[!th]
\centering
\includegraphics[width=\linewidth]{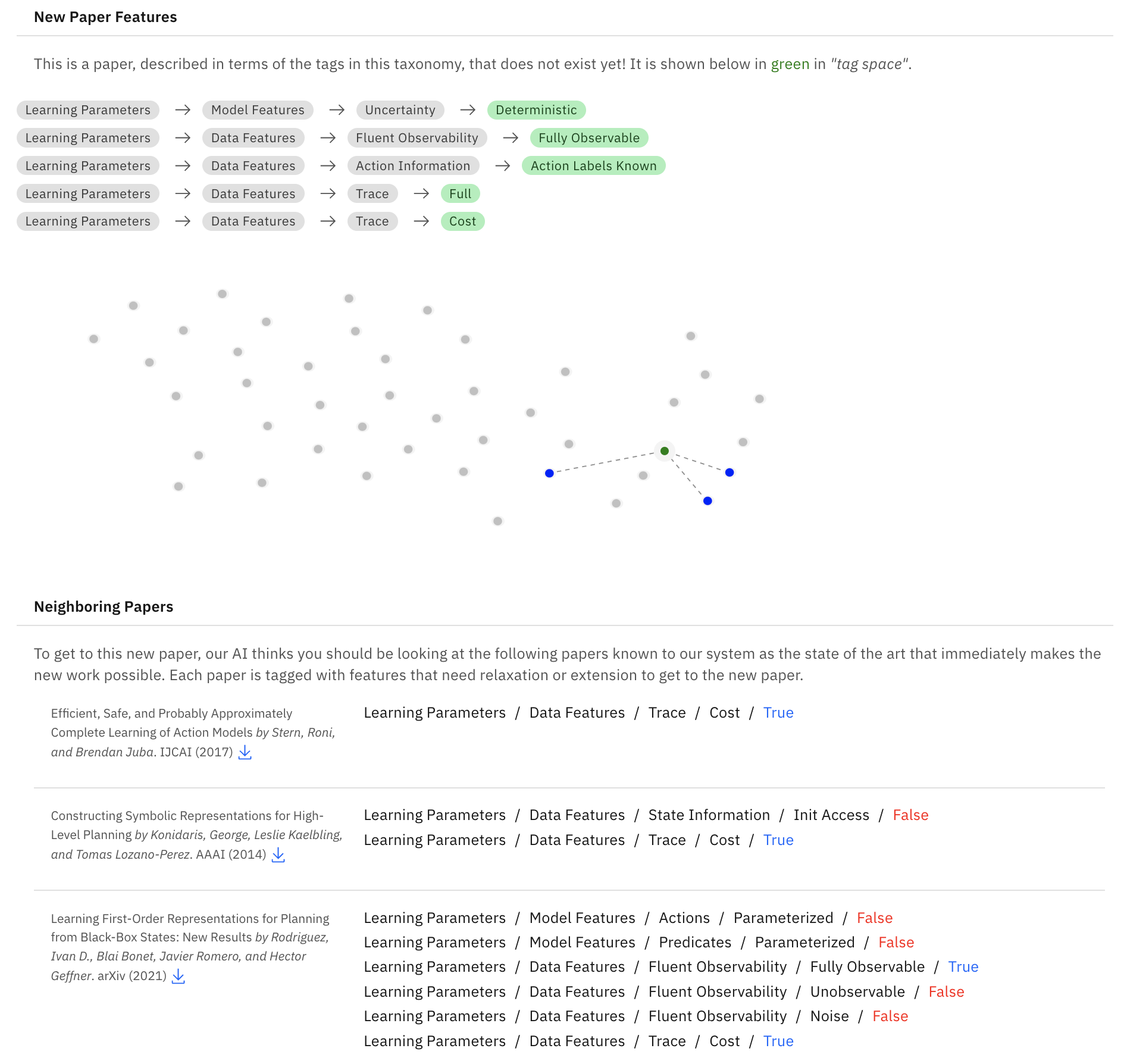}
\caption{Imagining future KEPS papers on the \macq visual interface.}
\label{fig:recommendation}
\end{figure*}

\subsection{Logical Encoding of Research Potential} 

Given the conjunction of constraints on the features, and the negation of the disjunction on the pre-existing literature (thus ruling out existing feature profiles), we have a logical theory where a satisfying assignment corresponds to a valid selection of features/assumptions about a model acquisition technique, and further is one that has not yet been explored in the known literature. Further preferences on specific features (e.g., wanting to only handle settings without parameters) can be included as unit clauses to further constrain the space of satisfying research configurations.

While this is a seemingly simple concept, there is tremendous potential in taking this viewpoint. We have implemented the above encoding and found that modern SAT solvers and knowledge compilers are readily capable of handling the theory. We further developed a novel research recommendation procedure for the space of model acquisition techniques. Making use of a knowledge compiler and repeated logical conditioning, the procedure is as follows:

\begin{enumerate}

\item Encode the constraints and existing techniques into a logical theory $T$.

\item Define a full set of soft preferences $P$ over the features that stipulate ``simple'' or ``nominal'' settings (e.g., fully observable over partially observable).

\item Run a full knowledge compilation on $T$, to get a d-DNNF representing all possible solutions $S$.

\item Iterate over $p \in P$, and if $p$ is consistent with $S$, enforce $p$ by setting $S = S \wedge p$.

\end{enumerate}

At the end of this procedure, we will have a single assignment to the full set of features that (1) adheres to a maximal number of preferences; and (2) differs from every other existing approach. Different orders for step 2 will potentially result in different final outcomes, and we will see one such example later on in this paper.

Finally, with a candidate area of unexplored research in hand, we can perform a matching algorithm to find the closest existing approaches to the one being proposed. Our system limits this to 3 and displays the core differences between the existing work and the newly proposed one. An example of this functionality is also provided below.

\subsection{Unwritten Paper Recommendations}

Figure \ref{fig:recommendation} illustrates this process 
in action on the \macq visual interface, introduced 
in Section \ref{subsec:viz}.
The visualization unfolds in three sections: 

\begin{itemize}
\item[-] The first part of the exposition describes the features of this newly imagined paper in terms of its features. The tag hierarchy is displayed.
\item[-] The hypothetical paper is now visualized in feature space: this view
shows where it belongs when all the papers known to \macq are projected
onto a latent space only\footnote{This view is slightly different from
the similarity view described Section \ref{subsec:viz}. There,
a document includes these features but also all the rest of the 
paper metadata in terms of authors, title, abstract, venue, and so on.} consisting of the features from Section \ref{subsec:features}.
\item[-] Finally, and perhaps most interestingly, the above visual leads into
neighbouring (in feature space) papers that the user can tap into as the state of the art closest to this new imagined paper. In addition to the metadata of
the neighbouring papers, \macq surfaces the features of those neighbours that
need to change (either relaxed or extended) in order to make a hop from 
a known relevant paper to this non-existent paper.
\end{itemize}

Currently, we are working on making this interface more interactive
so the user can query the system with a partial selection of 
papers and features of interest; and iteratively make hops to the
next imagined paper. With this feature implemented, we intend to do pilot
studies on how quickly we can onboard new students into a field using
this exposition and exploration technique over survey data.

    % \clearpage
% \section{Discussion}
% \label{sec:discussion}

\section{Open Research Questions}
\label{sec:open-questions}
Given the holistic analysis of the field that \macq offers, we have identified several promising areas for further research. Here, we highlight just a few of them.

\subsection{Operationalizing \macq}
\label{subsec:git}

Most of the existing approaches to model learning 
deal with a one-time model learning task while not taking into
account the operational considerations of deploying a system
with that model.
In reality, models are deployed and maintained over time. 
As such, such models drift \cite{bryce2016maintaining} and thus
systems require a certain level of hand-holding in terms of 
how to deal with such evolution of models they are deployed with.
For learning systems, this is increasingly becoming a trend \cite{gitlab},
with approaches that try to reconcile new models with past 
decisions \cite{bansal2019case} thereby offering a certain level
of consistency. We envisage similar processes to dovetail 
with the core \macq functionality when systems are deployed on top
of model learning tools in their portfolio.

\subsection{XAIP Crossover}
\label{subsec:xaip}

Interestingly, most of the models learned from data are
underdetermined -- i.e. there are many equivalent models that
can ``explain'' a set of observed circumstances.
This also holds for iterative or ``online'' approaches with 
a domain writer in the loop. 
In fact, some approaches e.g. \cite{yang2007learning} specifically 
looked at pattern mining tools to bias the learning approaches towards 
more likely models.
Even so, the decisions made by the model learning algorithm remain rather opaque
and it may well be the case that some models rejected by it could have made 
more sense when presented to the domain writer.
We are currently exploring the possible adoption of model-space reasoning 
techniques from the emerging field of explainable AI planning or XAIP
\cite{chakraborti-ijcai-2020} to engage in a more transparent model 
learning interface where the domain writer can be empowered to query 
and explore the trade-offs made among the equivalence class of 
models that satisfy a set of observed behaviours.

\subsection{Will AI write the papers of the future?}
\label{subsec:yolanda}

In her presidential address \cite{gil2022will} at AAAI 2020, 
Yolanda Gil, one of the early pioneers \cite{carbonell1990learning, gil1994learning}
in the field of model learning for automated planning,
asked: {\em ``Will AI write scientific papers in the future?''}.
The question was posed to facilitate 
an exploration of the influence that AI algorithms, 
from process management to knowledge discovery, increasingly 
have on our scientific endeavours. 
As we demonstrated in Section \ref{subsec:viz}, 
the set of exploratory features made available by \macq also belongs to this
emerging theme of collaboration between AI and the scientist: not to synthesize the papers directly, but rather to provide the automated insights necessary for researchers to know where next to look.
To the extent that that question applies to the KEPS community, 
\macq is most certainly going 
to (help) write the papers of the future!

% We welcome the 
% reader to have a gander at: \href{macq.com/explore}{macq.com/explore}.

    % \clearpage
    % \bibliographystyle{aaai21}
    \bibliography{references}

\end{document}